\documentclass[10pt, conference]{IEEEtran}
\IEEEoverridecommandlockouts
\usepackage{chngcntr}
\usepackage{graphicx}
\usepackage{subfig}
\usepackage{xcolor}
\usepackage{amsmath}
\usepackage{amssymb}
\usepackage{nccmath}
\usepackage{booktabs}
\usepackage{multirow}
\usepackage{bbm}
\usepackage{lipsum}
\usepackage{mathtools}
\usepackage{tikz}
\usetikzlibrary{positioning}
\usepackage{caption}
\usepackage{cite}
\usepackage{url}
\usepackage{eso-pic}
\def\BibTeX{{\rm B\kern-.05em{\sc i\kern-.025em b}\kern-.08em
    T\kern-.1667em\lower.7ex\hbox{E}\kern-.125emX}}

\begin{document}
\AddToShipoutPictureBG*{%
  \AtPageUpperLeft{%
    \hspace{0.98\paperwidth}%
    \raisebox{-\baselineskip}{%
      \makebox[0pt][r]{IEEE SoutheastCon2020}
}}}%
\tikzset{%
  every neuron/.style={
    circle,
    draw,
    minimum size=0.7cm
  },
  neuron missing/.style={
    draw=none, 
    scale=2,
    text height=0.233cm,
    execute at begin node=\color{black}$\vdots$
  },
}

\title{Solving The Lunar Lander Problem under Uncertainty using Reinforcement Learning\\
\thanks{978-1-7281-6861-6/20/\$31.00 ©2020 IEEE}
}

\author{\IEEEauthorblockN{Soham Gadgil}
\IEEEauthorblockA{\textit{Computer Science Department} \\
\textit{Stanford University}\\
\text{sgadgil@stanford.edu}
}
\and
\IEEEauthorblockN{Yunfeng Xin}
\IEEEauthorblockA{\textit{Electrical Engineering Department} \\
\textit{Stanford University}\\
\text{yunfeng.xin@stanford.edu}
\and
\IEEEauthorblockN{Chengzhe Xu}
\IEEEauthorblockA{\textit{Electrical Engineering Department} \\
\textit{Stanford University}\\
\text{czxu@stanford.edu}}
}}
\maketitle

\begin{abstract}
Reinforcement Learning (RL) is an area of machine learning concerned with enabling an agent to  navigate an environment with uncertainty in order to maximize some notion of cumulative long-term reward. In this paper, we implement and analyze two different RL techniques, Sarsa and Deep Q-Learning, on OpenAI Gym's \textit{LunarLander-v2} environment. We then introduce additional uncertainty to the original problem to test the robustness of the mentioned techniques. With our best models, we are able to achieve average rewards of 170+ with the Sarsa agent and 200+ with the Deep Q-Learning agent on the original problem. We also show that these techniques are able to overcome the additional uncertainties and achieve positive average rewards of 100+ with both agents. We then perform a comparative analysis of the two techniques to conclude which agent performs better.
\end{abstract}


\section{Introduction}
Over the past several years, reinforcement learning \cite{sutton2018reinforcement} has been proven to have a wide variety of successful applications including robotic control \cite{kober2013reinforcement, riedmiller2009reinforcement}. Different approaches have been proposed and implemented to solve such problems \cite{kaelbling1996reinforcement, szepesvari2010algorithms}. In this paper, we solve a well-known robotic control problem --  the lunar lander problem -- using different reinforcement learning algorithms, and then test the agents' performances under environment uncertainties and agent uncertainties. The problem is interesting since it presents a simplified version of the task of landing optimally on lunar surface, which in itself has been a topic of extensive research \cite{brady2010challenge, cho2009optimal, liu2008optimal}. The added uncertainties are meant to model the situations that a real spacecraft might face while landing on the lunar surface.

\section{Problem Definition}
We aim to solve the lunar lander environment in the OpenAI gym kit using reinforcement learning methods.\footnote{Our code is available at \url{https://github.com/rogerxcn/lunar_lander_project}} The environment simulates the situation where a lander needs to land at a specific location under low-gravity conditions, and has a well-defined physics engine implemented.

The main goal of the game is to direct the agent to the landing pad as softly and fuel-efficiently as possible. The state space is continuous as in real physics, but the action space is discrete.

\section{Related Work}
There has been previous work done in solving the lunar lander environment using different techniques. \cite{8882916} makes use of modified policy gradient techniques for evolving neural network topologies. \cite{lu2019controlmodelbased} uses a control-model-based approach that learns the optimal control parameters instead of the dynamics of the system. \cite{FLAIRS1918312} explores spiking neural networks as a solution to OpenAI virtual environments. 

These approaches show the effectiveness of a particular algorithm for solving the problem. However, they do not consider additional uncertainty. Thus, we aim to first solve the lunar lander problem using traditional Q-learning techniques, and then analyze different techniques for solving the problem and also verify the robustness of these techniques as additional uncertainty is added. 

\begin{figure}[!t]
    \centerline{
    \includegraphics[width=2.0in]{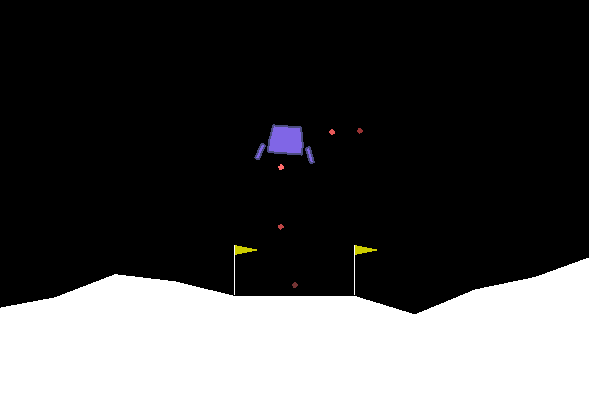}}
    \caption{Visualization of the lunar lander problem.}
    \label{fig:vis_game}
\end{figure}

\section{Model}
\subsection{Framework}
The framework used for the lunar lander problem is gym, a toolkit made by OpenAI \cite{brockman2016openai} for developing and comparing reinforcement learning algorithms. It supports problems for various learning environments, ranging from Atari games to robotics. The simulator we use is called \textit{Box2D} and the environment is called \textit{LunarLander-v2}.
\subsection{Observations and State Space}
The observation space determines various attributes about the lander. Specifically, there are 8 state variables associated with the state space, as shown below:

    \begin{align*}
        state\rightarrow\left\{
        \begin{array}{ll}
            \text{$x$ coordinate of the lander}   \\
             \text{$y$ coordinate of the lander}  \\
            \text{$v_x$, the horizontal velocity} \\
            \text{$v_y$, the vertical velocity}  \\
            \text{$\theta$, the orientation in space} \\
            \text{$v_{\theta}$, the angular velocity}\\
            \text{Left leg touching the ground (Boolean)}\\
            \text{Right leg touching the ground (Boolean)}
        \end{array}
        \right.
    \end{align*}

All the coordinate values are given relative to the landing pad instead of the lower left corner of the window. The $x$ coordinate of the lander is $0$ when the lander is on the line connecting the center of the landing pad to the top of the screen. Therefore, it is positive on the right side of the screen and negative on the left. The $y$ coordinate is positive at the level above the landing pad and is negative at the level below.

\subsection{Action Space}
There are four discrete actions available: do nothing, fire left orientation engine, fire right orientation engine, and fire main engine. Firing the left and right engines introduces a torque on the lander, which causes it to rotate, and makes stabilizing difficult.

\subsection{Reward}
Defining a proper reward directly affects the performance of the agent. The agent needs to maintain both a good posture mid-air and reach the landing pad as quickly as possible. Specifically, in our model, the reward is defined to be:

\begin{equation}
\begin{split}
    \text{Reward}(s_t) = -100* (d_t - d_{t-1}) - 100* (v_t - v_{t-1})\\ -100* (\omega_t - \omega_{t-1}) + \text{ hasLanded}(s_t)
\end{split} 
\end{equation}

where $d_t$ is the distance to the landing pad, $v_t$ is the velocity of the agent, and $\omega_t$ is the angular velocity of the agent at time $t$. $hasLanded()$ is the reward function of landing, which is a linear combination of the boolean state values representing whether the agent has landed softly on the landing pad and whether the lander loses contact with the pad on landing. 

With this reward function, we encourage the agent to lower its distance to the landing pad, decrease the speed to land smoothly, keep the angular speed at minimum to prevent rolling, and not to take off again after landing.

\begin{figure}[!t]
    \centering
    \includegraphics[width=2.8in]{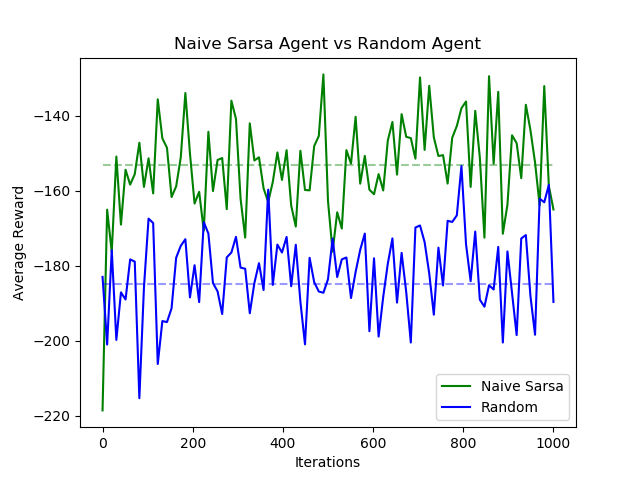}
    \caption{Performance comparison between Sarsa agent under naive state discretization and random agent.}
    \label{fig:naivesarsavsrandom}
\end{figure}

\section{Approaches}
\subsection{Sarsa}
Since we do not encode any prior knowledge about the outside world into the agent and the state transition function is hard to model, Sarsa \cite{kochenderfer2015decision} seems to be a reasonable approach to train the agent using an exploration policy. The update rule for Sarsa is:

\begin{multline}
    Q(s_t, a_t) = Q(s_t, a_t) + \\
\alpha [r_t + Q(s_{t+1}, a_{t+1}) - Q(s_t, a_t)]
\end{multline}

At any given state, the agent chooses the action with the highest Q value corresponding to:

\begin{equation}
\begin{aligned}
    a = argmax_{a \in actions} Q(s, a)
\end{aligned} 
\end{equation}

From the equation we can see that we need to discretize the states and assign Q values for each state-action pair, and that we also need to assign a policy to balance exploitation and exploration since Sarsa is an on-policy algorithm.

Intuitively, a simple exploration policy can be the $\epsilon$-greedy policy \cite{RodriguesGomes:2009:DAM:1553374.1553422}, where the agent randomly chooses an action with probability $\epsilon$ and chooses the best actions with probability $1 - \epsilon$. A simple way of discretizing the state is to divide each continuous state variable into several discrete values. However, as shown in Fig. \ref{fig:naivesarsavsrandom}, the result shows that the agent can only reach marginally better performance than a random agent, and cannot manage to get positive rewards even after 10,000 episodes of training. It then becomes obvious that we cannot simply adopt these algorithms out-of-the-box, and we need to tailor them for our lunar lander problem.

\begin{figure}[!t]
    \centering
    \includegraphics[width=2.5in]{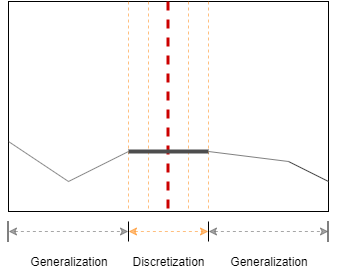}
    \caption{Diagram showing the state discretization and generalization scheme of the x coordinate state variable.}
    \label{fig:discretization}
\end{figure}

\subsubsection{State Discretization}
There are six continuous state variables and two boolean state variables, so the complexity of the state space is on the order of $O(n^6 \times 2 \times 2)$, where $n$ is the number of discretized values for each state variable. Thus, even discretizing each state variable into 10 values can produce 400,000 states, which is far too large to explore within a reasonable number of training episodes. It explains the poor performance we were seeing previously. Therefore, we need to devise a new discretization scheme for these states.

Specifically, we examine the initial Q values learned by the agent and observe that the agent wants to move to the right when it is in the left half of the space, and move to the left when it is in the right half of the space. As a result, all the x coordinates far from the center can be generalized into one single state because the agent will always tend to move in one direction (as shown in Fig. \ref{fig:discretization}), which helps reduce the state space.

Therefore, we define the discretization of the x coordinate at any state with the discretization step size set at 0.05 as:
\begin{equation}
\begin{aligned}
    d(x) = \min(\lfloor \frac{n}{2} \rfloor, \max(-\lfloor \frac{n}{2} \rfloor, \frac{x}{\text{0.05}}))
\end{aligned} 
\end{equation}

The same intuition of generalization is applied to other state variables as well. In total, we experiment with 4 different ways of discretizing and generalizing the states with different number of discretization levels.

\subsubsection{Exploration Policy}
As the size of the state space exceeds 10,000 even after the process of discretization and generalization, the probability $\epsilon$ in the $\epsilon$-greedy policy needs to be set to a fairly large number ($\sim$20\%) for the agent to explore most of the states within a reasonable number of episodes. However, this means that the agent will pick a sub-optimal move once every five steps.

In order to minimize the performance impact of the $\epsilon$-greedy policy while still getting reasonable exploration, we decay $\epsilon$ in different stages of training. The intuition is that the agent in the initial episodes knows little about the environment, and thus more exploration is needed. After an extensive period of exploration, the agent has learned enough information about the outside world, and needs to switch to exploitation so that the Q value for each state-action pair can eventually converge.

Specifically, the epsilon is set based on the following rules:

    \begin{align*}
        \epsilon=\left\{
        \begin{array}{ll}
            0.5 & \text{\#Iteration} \in [0, 100) \\
            0.2 & \text{\#Iteration} \in [100, 500) \\
            0.1 & \text{\#Iteration} \in [500, 2500) \\
            0.01 & \text{\#Iteration} \in [2500, 7500) \\
            0 & \text{\#Iteration} \in [7500, 10000)
        \end{array}
        \right.
    \end{align*}

\subsection{Deep Q-Learning}
Since there is model uncertainty in the problem, Q-learning is another approach which can be used to solve the environment. For this problem, we use a modified version of Q-learning, called deep Q-learning (DQN) \cite{mnih2013playing, mnih2015human} to account for the continuous state space. The DQN method makes use of a multi-layer perceptron, called a Deep Q-Network (DQN), to estimate the Q values. The input to the network is the current state (8-dimensional in our case) and the outputs are the Q values for all state-action pairs for that state. The Q-Learning update rule is as follows:
\begin{equation}    
Q(s,a) = Q(s, a) + \alpha(r + \gamma \max\limits_{a^{\prime}}Q(s^{\prime}, a^{\prime}) - Q(s, a))
\end{equation}
The optimal $Q$-value $Q^{*}(s,a)$ is estimated using the neural network with parameters $\theta$. This becomes a regression task, so the loss function at iteration $i$ is obtained by the temporal difference error:

\begin{align}
\mathcal{L}_i(\theta_i)&=\mathbbm{E}_{(s,a)\sim\rho(.)}[(y_i-Q(s,a;\theta_i))^2]
\end{align}
where 
\begin{align}
    y_i=\mathbbm{E}_{s^{\prime}\sim\mathcal{E}}[r + \gamma \max\limits_{a^{\prime}}Q(s^{\prime}, a^{\prime}; \theta_{i-1})]
 \end{align}

Here, $\theta_{i-1}$ are the network parameters from the previous iteration, $\rho(s,a)$ is a probability distribution over states $s$ and actions $a$, and $\mathcal{E}$ is the environment the agent interacts with.  Gradient descent is then used to optimise the loss function and update the neural network weights. The neural network parameters from the previous iteration, $\theta_{i-1}$, are kept fixed while optimizing $\mathcal{L}_i(\theta_i)$. Since the next action is selected based on the greedy policy, Q-learning is an off policy algorithm \cite{watkins1992q}.

One of the challenges here is that the successive samples are highly correlated since the next state depends on the current state and action. This is not the case in traditional supervised learning problems where the successive samples are i.i.d. To tackle this problem, the transitions encountered by the agent are stored in a replay memory $\mathcal{D}$. Random minibatches of transitions $\{s,a,r,s^{\prime}\}$ are sampled from $\mathcal{D}$ during training to train the network. This technique is called \textit{experience replay} \cite{lin1993reinforcement}.

\begin{figure}[!t]
\centering
\begin{tikzpicture}[x=0.85cm, y=0.85cm, >=stealth]

\foreach \m/\l [count=\y] in {1,2,3,missing,4}
  \node [every neuron/.try, neuron \m/.try] (input-\m) at (0,2.5-\y) {};

\foreach \m [count=\y] in {1,missing,2}
  \node [every neuron/.try, neuron \m/.try ] (hidden_1-\m) at (2,2-\y*1.25) {};
  
 \foreach \m [count=\y] in {1,missing,2}
  \node [every neuron/.try, neuron \m/.try ] (hidden_2-\m) at (4,2-\y*1.25) {};

\foreach \m [count=\y] in {1,missing,2}
  \node [every neuron/.try, neuron \m/.try ] (output-\m) at (6,1.5-\y) {};

\foreach \l [count=\i] in {1,2,3,8}
  \draw [<-] (input-\i) -- ++(-1,0)
    node [above, midway] {$s_\l$};

\foreach \l [count=\i] in {1, 128}
  \node [above] at (hidden_1-\i.north) {$h_{\l}$};

\foreach \l [count=\i] in {1,128}
  \node [above] at (hidden_2-\i.north) {$h_{\l}$};

\foreach \l [count=\i] in {1,4}
  \draw [->] (output-\i) -- ++(1,0)
    node [above, midway] {$a_\l$};

\foreach \i in {1,...,4}
  \foreach \j in {1,...,2}
    \draw [->] (input-\i) -- (hidden_1-\j);

\foreach \i in {1,...,2}
  \foreach \j in {1,...,2}
    \draw [->] (hidden_1-\i) -- (hidden_2-\j);

\foreach \i in {1,...,2}
  \foreach \j in {1,...,2}
    \draw [->] (hidden_2-\i) -- (output-\j);

\foreach \l [count=\x from 0] in {Input, Hidden, Hidden, Ouput}
  \node [align=center, above] at (\x*2,2) {\l \\ layer};

\end{tikzpicture}
\captionof{figure}{Neural Network used for Deep Q-Learning}
\label{fig:nn}
\end{figure}
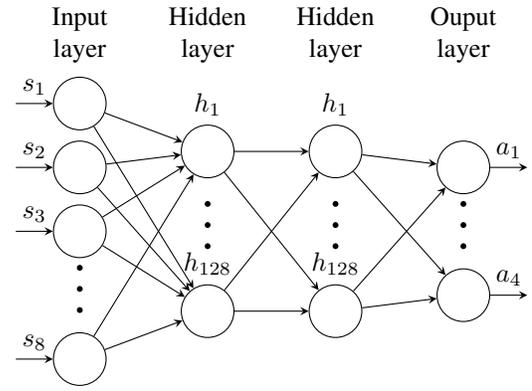

We use a 3 layer neural network, as shown in Fig. \ref{fig:nn}, with 128 neurons in the hidden layers. We use $ReLU$ activation for the hidden layers and $LINEAR$ activation for the output layer. The number of hidden neurons are chosen based on analyzing different values, as shown in section $6.2.1$.  
The learning rate used is $0.001$ and the minibatch size is $64$.
\subsubsection{Exploration Policy}
Similar to Sarsa, an improved $\epsilon$-greedy policy is used to select the action, with $\epsilon$ starting at 1 to favor exploration and decaying by $0.996$ for every iteration, until it reaches 0.01, after which it stays the same.

\section{Experiments}
\subsection{The Original Problem}
The goal of the problem is to direct the lander to the landing pad between two flag poles as softly and fuel efficiently as possible. Both legs of the lander should be in contact with the pad while landing. The lander should reach the landing pad as quickly as possible, while maintaining a stable posture and minimum angular speed. Also, once landed, the lander should not take off again. In the original problem, uncertainty is added by applying a randomized force to the center of the lander at the beginning of each iteration. This causes the lander to be pushed in a random direction. The lander must recover from this force and head to the landing pad.

We experiment with tackling the original problem using Sarsa and deep Q-learning as described in our approach section, and our observations are demonstrated in section $7$.

\subsection{Introducing Additional Uncertainty}
After solving the original lunar lander problem, we analyze how introducing additional uncertainty can affect the performance of the agents and evaluate their robustness to different uncertainties.
\subsubsection{Noisy Observations}
Retrieving the exact state of an object in a physical world can be hard, and we need to rely on noisy observations such as a radar to infer the real state of the object. Thus, instead of directly using the exact observations provided by the environment, we add a zero-mean Gaussian noise of scale 0.05 into our observation of the location of the lander. The standard deviation is deliberately set to 0.05, which corresponds to our discretization step size. Specifically, for each observation of $x$, we sample a random zero-mean Gaussian noise
\begin{equation}
    s \sim \mathcal{N}(\mu = 0, \sigma = 0.05)
\end{equation}
and add the noise to the observation, so that the resulting random variable becomes
\begin{equation}
    Observation(x) \sim \mathcal{N}(x, 0.05)
\end{equation}

We then evaluate the resulting performance of two agents: one using the original Q values from the original problem, and the other using the Q values trained under such noisy observations. 

We notice that we can frame this problem as a POMDP (Partially Observable Markov Decision Process), and compare its performance with the two Sarsa agents mentioned above. We calculate the alpha vector of each action using one-step lookahead policy using the Q values from the original problem, and calculate the belief vector using the Gaussian probability density function

\begin{equation}
    \text{PDF}(x) = \frac{1}{\sqrt{2\pi}\sigma} e^{-\frac{(o(x) - x)^2}{2\sigma^2}}
\end{equation}

This way, we can get the expected utility of each action under uncertainty by taking the inner product of the corresponding alpha vector and the belief vector. The resulting policy simply picks the action with the highest expected utility.

Notice that we could have used transition probabilities of locations to assist in determining the exact location of the agent. However, after experimenting with different transition probability functions, we concluded that the transition probability in a continuous space is very hard to model, and naive transition probability functions will cause the agent to perform even worse than the random agent. 

\begin{figure}[!t]
    \centering
    \includegraphics[width=2.8in]{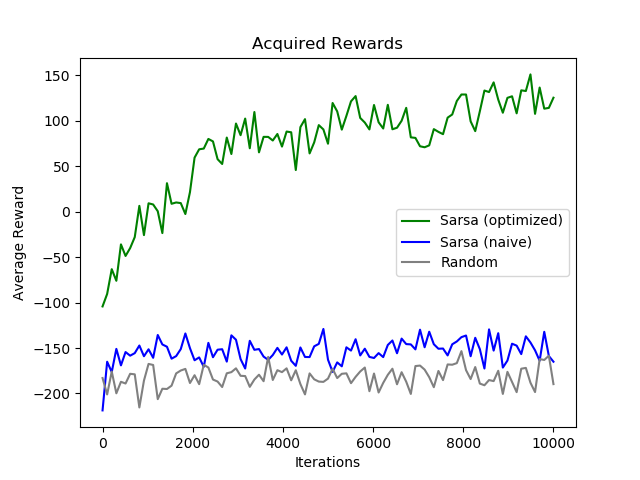}
    \caption{Performance comparison of our state discretization and policy scheme (green), naive state discretization and policy scheme (blue), and random agent (grey).}
    \label{fig:sarsavsnaivevsrandom}
\end{figure}

\subsubsection{Random Engine Failure}
Another source of uncertainty in the physical world can be random engine failures due to the various unpredictable conditions in the agent's environment. The model needs to be robust enough to overcome such failures without impacting performance too much. To simulate this, we introduce action failure in the lunar lander. The agent takes the action provided $80\%$ of the time, but $20\%$ of the time the engines fail and it takes no action even though the provided action is firing an engine. 

\subsubsection{Random Force}
 Uncertainty can also come from unstable particle winds in the space such as solar winds, which result in random forces being applied to the center of the lander while landing. The model is expected to manage the random force and have stable Q maps with enough fault tolerance. 
 
 We apply a random force each time the agent interacts with the environment and modify the state accordingly. The force is sampled from a Gaussian distribution for better simulation of real-world random forces. The mean and variance of the Gaussian distribution are set in proportion to the engine power to avoid making the random force either too small to have any effect on the agent or too large to maintain control of the agent. 
 
 \begin{figure}[!t]
    \centering
    \includegraphics[width=2.8in]{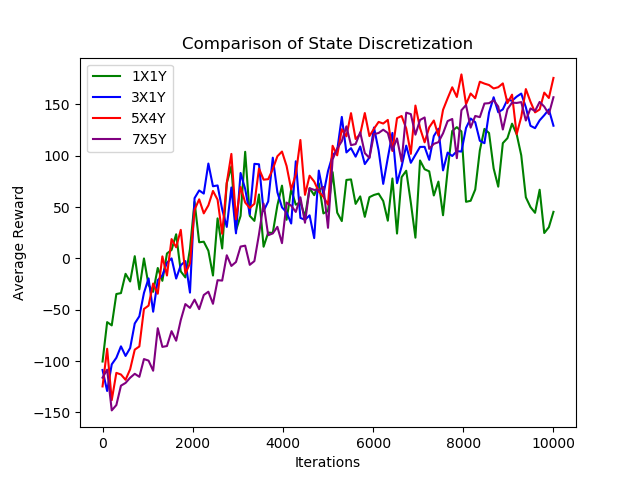}
    \caption{Performance comparison of different state discretization schemes in Sarsa.}
    \label{fig:sarsa_discretization}
\end{figure}

\section{Results and Analysis}
\subsection{Sarsa}
\subsubsection{The Original Problem}
Fig. \ref{fig:sarsavsnaivevsrandom} shows the average reward acquired (over the previous 10 episodes) by the the random agent and the Sarsa agent with naive state discretization and our customized discretization scheme. With a naive discretization which quantizes each state variable into 10 equal steps, the agent cannot learn about the outside world very effectively even after 10,000 episodes of training. Due to the huge state space, the acquired reward is only marginally better than the random agent. In contrast, with our customized discretization scheme which combines small-step discretization and large-step generalization, the agent is able to learn the Q values rapidly and gather positive rewards after 500 episodes of training. The results show that for continuous state spaces, proper discretization and generalization of the states helps the agent learn the values faster and better.

Fig. \ref{fig:sarsa_discretization} also shows how different discretization schemes affect the learning speed and the final performance the agent is able to achieve. The notation $aXbY$ is used to denote that the the x coordinate is discretized into $a$ levels while the y coordinate is discretized into $b$ levels. The results indicate that as the number of discretization levels increases, the agent in general learns the Q values more slowly, but is able to achieve higher performance after convergence. The discretization scheme 5X4Y does slightly better than the scheme 7X5Y, indicating that further discretization will not help.

\subsubsection{Handling Noisy Observations}
\begin{figure}[!t]
    \centering
    \includegraphics[width=2.8in]{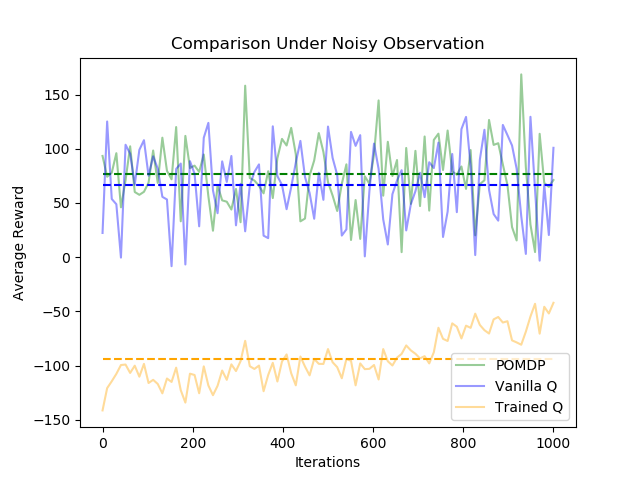}
    \caption{Comparison of different agent performance under noisy observations.}
    \label{fig:noisy_obs}
\end{figure}

Fig. \ref{fig:noisy_obs} shows the results after feeding the noisy observations under three agents. The first agent uses the Q values learned from the original problem under discretization scheme 5X4Y and takes the noisy observations as if they were exact. The second agent is re-trained under the noisy environment using the noisy observations. The third agent uses the Q values learned from the original problem, but uses one-step lookahead alpha vectors to calculate the expected reward for each action.

Each data point in Fig. \ref{fig:noisy_obs} represents the average acquired reward in 10 episodes and can help eliminate outliers. The results show that the POMDP agent (POMDP in Fig. \ref{fig:noisy_obs}) receives the highest average rewards and outperforms the other agents. Of the other two agents, the agent trained under noisy observations (Trained Q in Fig. \ref{fig:noisy_obs}) fails to generalize information from these episodes.

In general, there is a significant performance impact in terms of average acquired rewards with the added uncertainty of noisy observation, and such a result is expected: when the agent is close to the center of the space, a noisy x observation can significantly change the action which the agent picks. For example, when $x = 0.05$, a noisy observation has a 15.87\% probability of flipping the sign so that $x < 0$ according the Gaussian cumulative distribution function

\begin{equation}
    \text{CDF}(x) = \frac{1}{\sqrt{2\pi} \times 0.05} \int_{-\infty}^{-0.05} e^{-\frac{(x - 0.05)^2}{2 \times 0.05^2}} dx
\end{equation}

This means that the noise has a decent chance of tricking the agent into believing that it is in the left half of the screen while it is in fact in the right half of the screen. Therefore, the agent will pick an action that helps the agent move right, instead of original optimal action of moving left.

The POMDP agent has the correct learned Q value and takes the aforementioned sign-flipping observation scenario into account using the belief vector, which explains why it is performing the best by getting the highest average rewards. The agent trained under noisy observation, however, is learning the wrong Q value in the wrong state due to the noisy observation and does not take the noisy observation into account. Thus, it is performing the worst of all three agents.

\begin{figure}[!t]
    \centering
    \includegraphics[width=2.8in]{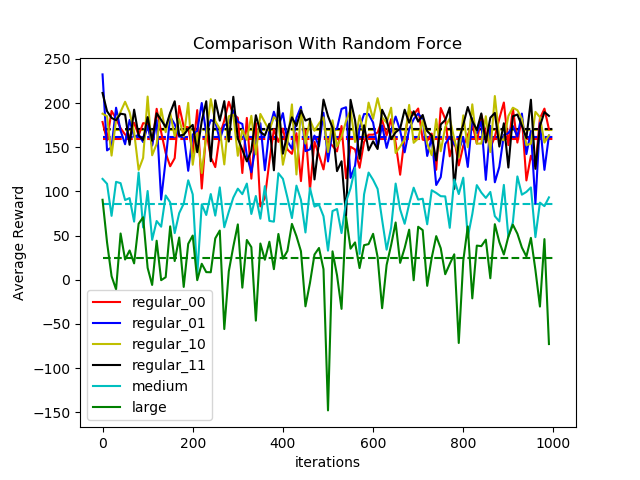}
    \caption{Comparison of Sarsa agent performance under different random forces.}
    \label{fig:Force_result}
\end{figure}

\subsubsection{Handling Random Force} \label{random_force}
Fig. \ref{fig:Force_result} shows the result after applying different random forces to the Sarsa agent under state discretization of 5X4Y. 

In the experiments, we introduce three kinds of random forces: regular, medium and large. In the regular cases, we ensure that random forces do not go too much beyond the agent's engine power. In the medium case, we relax that constraint, and in the large case, we ensure that the agent cannot control itself because the random force becomes much larger than the engine power. The details are described as follows:

1) regular$\_$00: mean equals 0 and variance equals $engine\_power/3$

2) regular$\_$01: mean equals 0 on the x-axis and  $engine\_power/6$ on the y-axis, variance equals $engine\_power/3$ 

3) regular$\_$10: mean equals $engine\_power/6$ on the x-axis and 0 on the y-axis, variance equals $engine\_power/3$ 

4) regular$\_$11: mean equals $engine\_power/6$ on both x-axis and y-axis, variance equals $engine\_power/3$ 

5) medium: mean equals $engine\_power$ on both x-axis and y-axis, variance equals $engine\_power*3$ 

6) large: mean equals $engine\_power*2$  on both x-axis and y-axis, variance equals $engine\_power*5$

The result suggests that agents would perform well and offset the effect of the random force in regular cases, while in the medium and large cases where random forces are more likely to exceed the maximum range engines could compensate, there would be an obvious reduction in reward indicating that the random forces make landing harder. The results reveal the fact that Sarsa agents have learned a robust and smooth Q map where similar state-action pairs would have similar Q value distributions. Slight state variations caused by random forces would have small influences on Q value and action selection, which increases the fault tolerance of the agent.

When state variations become too large, Q maps would be noisy and total rewards would decrease, in which case agents would tend to lose control because of the random forces.

\begin{figure}[!t]
    \centering
    \includegraphics[width=2.8in]{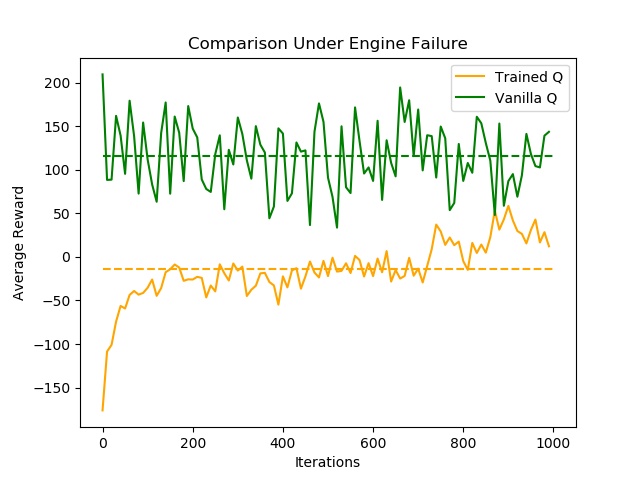}
    \caption{Comparison of Sarsa agent performance under engine failure}
    \label{fig:sarsa_engine_fail}
\end{figure}
\subsubsection{Handling Random Engine Failure}
Fig. \ref{fig:sarsa_engine_fail} shows the results after adding engine failure to the original environment. The first agent uses Q values learned from the original problem under discretization scheme 5X4Y and the second agent is retrained in the environment with engine failure. The agent reusing Q values (Vanilla Q in Fig. \ref{fig:sarsa_engine_fail}) does well and is able to achieve positive average rewards of 100+. This shows that the Sarsa agent using the discretization scheme is robust enough to recover from sudden failures in the engines. The retrained Sarsa agent (Trained Q in Fig. \ref{fig:sarsa_engine_fail}) is not able to achieve rewards as high as the vanilla Q agent, indicating that the agent is not able to generalize information from these episodes. 

Again, there is a performance drop with the added uncertainty of engine failure. This is understandable, since an engine failure causes the lander to behave unexpectedly and the effect gets compounded as more engine failures take place. 

\subsection{Deep Q-Learning}
\subsubsection{The Original Problem}
\begin{figure}[!t]
    \centering
    \includegraphics[width=2.8in]{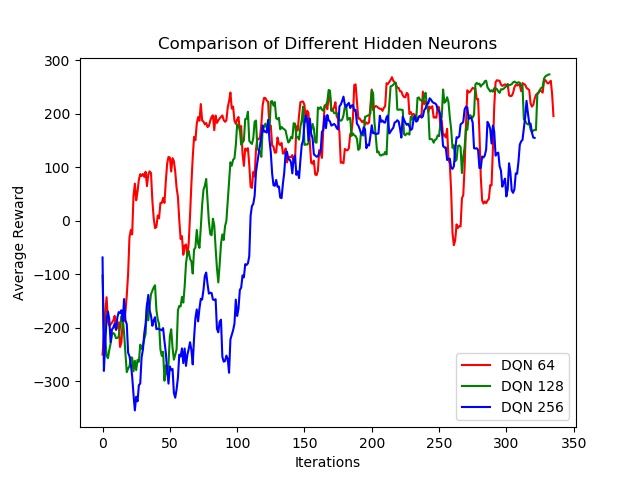}
    \caption{Average reward obtained by DQN agent for different number of hidden neurons}
    \label{fig:dqn_agent}
\end{figure}

Fig. \ref{fig:dqn_agent} shows the average reward obtained (over the previous 10 episodes) by the DQN agent for different number of neurons in the hidden layers of the neural network used for predicting the Q-values. At the start, the average reward is negative since the agent is essentially exploring all the time and collecting more transitions for storing in memory $\mathcal{D}$. As the agent learns, the Q-values start to converge and the agent is able to get positive average reward. For both the plots, the DQN agent is able to converge pretty quickly, doing so in $\sim$ 320 iterations. The DQN implementation performs well and results in good scores. However, the training time for the DQN agent is quite long ($\sim8$ hours). It is observed that the neural network with $128$ neurons in the hidden layers  reaches the highest converged average rewards, and is chosen as the hyperparameter value.

\begin{figure}[!t]
    \centering
    \includegraphics[width=2.8in]{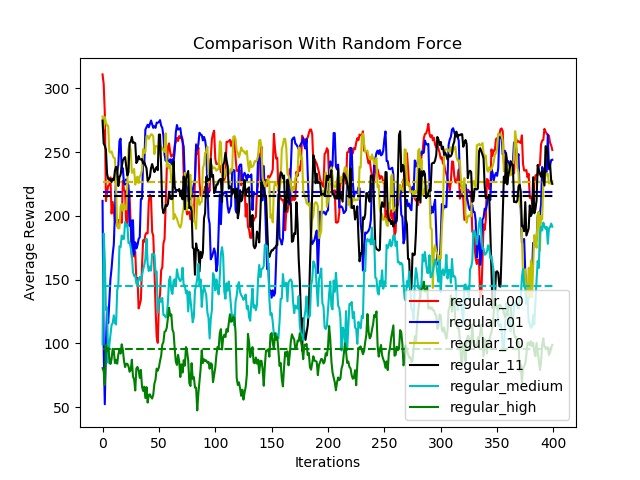}
    \caption{Comparison of DQN agent performance under different random forces}
    \label{fig:dqn_agent_const_force}
\end{figure}

\subsubsection{Handling Random Force}
Fig. \ref{fig:dqn_agent_const_force} shows the result after applying different random forces described in section \ref{random_force} to the DQN agent. The result looks similar to the one obtained with the Sarsa agent. The agent performs well in the the regular cases and the obtained reward curves are very similar. In the the medium and the large cases, the average reward drops and the agent performs the worst for the large case, and is barely able to get rewards above 100. This suggests that the agent is able to adapt well to slight deviations in the state of the lander due to the random forces and the rewards obatined are almost equal to the ones obtained in the original environment. However, when the random forces become too large, the agent is not able to overcome the effect of the forces to land optimally, and there is a reduction in the rewards obtained.

\begin{figure}[!t]
    \centering
    \includegraphics[width=2.8in]{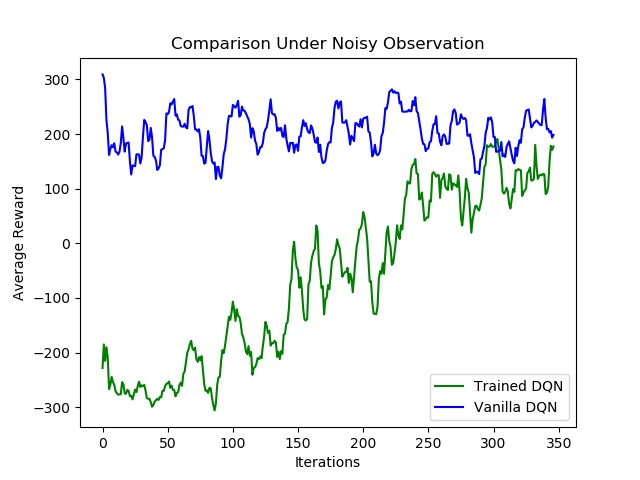}
    \caption{Comparison of DQN agent performance under noisy observations}
    \label{fig:dqn_agent_noisy_obs}
\end{figure}

\subsubsection{Handling Noisy Observations}
Fig. \ref{fig:dqn_agent_noisy_obs} shows a comparison between the agent using Q-values learned from the original problem (Vanilla DQN) and the retrained DQN agent with noisy observations (Trained DQN). The DQN agent trained under noisy observations is able to obtain positive rewards but fails to perform as well as the vanilla DQN agent. This can also be because a noisy x observation close to the center of the environment can significantly affect the action chosen by the agent. However, unlike the Sarsa agent which fails to generalize any information, the retrained DQN agent is able to capture some information about the environment even with noisy observations, which can be seen by the upward trend of the reward curve. 

\subsubsection{Handling Random Engine Failure}
\begin{figure}[!t]
    \centering
    \includegraphics[width=2.8in]{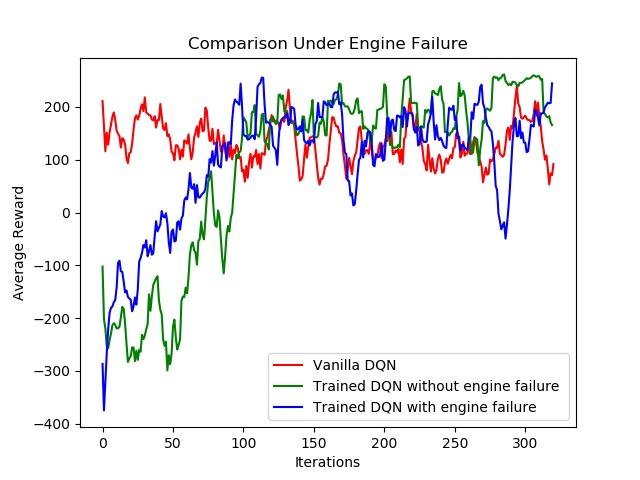}
    \caption{Comparison of DQN agents under engine failure}
    \label{fig:dqn_agent_engine_failure}
\end{figure}

Fig. \ref{fig:dqn_agent_engine_failure} shows a comparison between the agent using Q-values learned from the original problem (Vanilla DQN), the re-trained DQN agent when there are engine failures, and the original DQN agent when there are no engine failures. For all the plots, the number of neurons in the hidden layers of the neural network is 128. The vanilla DQN agent performs well on the new environment with random engine failures and is able to obtain positive average rewards of $100+$. This shows that even without retraining, the original agent is able to adjust to the uncertainty.

For the re-trained agents, at the start, the curves are similar since this is the exploration phase. Both agents are taking random actions and engine failure does not affect the reward much. However, in the later iterations, as the agents learn the environment, the lander without engine failure achieves higher average reward and is less erratic as compared to the lander with engine failure. This is expected since the random engine failures require the agent to  adjust to the unexpected behavior while trying to maximise reward. However, even with engine failure, the agent shows the same increasing trend in average reward as the original problem and is able to achieve positive rewards around $100$. Also, both the retrained agents achieve higher average rewards than the vanilla DQN. This shows that the DQN agent is able to estimate the optimal Q-values well even with the added uncertainty. 

\subsection{Comparative Analysis}
Based on the results of testing the agents in different environments, we can observe that both the Sarsa and DQN agents are able to achieve rewards of 100+ even in cases of added uncertainties, except for the retrained Sarsa agent under noisy observations, in which case the POMDP agent performs well. This shows that both the methods work well in practice and are robust enough to handle environments with added uncertainty. 

Comparing the results of the two methods, it can be observed that the DQN agent with a 3-layer neural network of 128 hidden neurons consistently gets higher average rewards, both for the original problem and for the problems with added uncertainty, than the Sarsa agent under the 5X4Y discretization scheme. This can be because with the Sarsa agent, we lose information about the state on discretization, which can affect how well the agent learns the environment. The 
DQN agent doesn't discretize the state space and uses all the information that the environment provides. Also, Q-learning is an off policy algorithm in which it learns the optimal policy using the absolute greedy policy by selecting the next action which maximises the Q-value. Since this is a simulation, the agent's performance during the training process doesn't matter. In such situations, the DQN agent would perform better since it learns an optimal greedy policy which we switch to eventually.

However, the DQN agent seems to be more erratic than the Sarsa agent, especially in the environment with noisy observations. There are drops in the acquired average reward for both the agents, which can be because of the randomness associated with the original environment and the added uncertainties. These drops are more frequent in the DQN agent than the Sarsa agent, which shows that even though the DQN agent is able to achieve higher average rewards, it is not as stable as the Sarsa agent. Also, the DQN agent takes twice as long to train as the Sarsa agent.

\section*{Conclusion}
In conclusion, we observe that both the Sarsa and the DQN agents perform well on the orignal lunar lander problem. When we introduce additional uncertainty, both agents are able to adapt to the new environments and achieve positive rewards. However, the re-trained Sarsa agent fails to handle noisy observations. This is understandable since the noisy observations affect the underlying state space and the agent isn't able to generalize information from its environment during training. The POMDP agent performs well with noisy observations and is able to get positive average rewards since it makes use of belief vectors to model a distribution over the possible next states. Overall, the DQN agent performs better than the Sarsa agent.

For future work, we would like to combine the different uncertainties together and analyze how the different agents perform. This will provide a more holistic overview of the robustness of different agents.

\bibliographystyle{./IEEEtran}
\bibliography{./references}

\end{document}